\documentclass[10pt, a4paper]{article}

\usepackage[final]{lrec-coling2024} 
\usepackage[T1]{fontenc}
\usepackage{subcaption}
\usepackage{textcomp}
\usepackage{amsmath}
\usepackage{amssymb}

\DeclareTextFontCommand{\textttenglish}{\ttenglishfamily}

\title{BenLLM-Eval: A Comprehensive Evaluation into the Potentials and Pitfalls of Large Language Models on Bengali NLP} 

\name{Mohsinul Kabir$^{\text{1},\text{*}}$\thanks{$^{\text{\textbf{*}}}$The first two authors contributed equally to this work.}, Mohammed Saidul Islam$^{\text{2},{\text{*},\text{\textdagger}}}$,\thanks{$^{\text{\textdagger}}$Corresponding Authors.} Md Tahmid Rahman Laskar$^{\text{2},\text{5},\text{\textdagger}}$,
\\ {\bf \large Mir Tafseer Nayeem$^{\text{3}}$,} {\bf \large M Saiful Bari$^{\text{4}}$,} {\bf \large Enamul Hoque$^{\text{2}}$}}

\address{$^{1}$Islamic University of Technology, $^{2}$York University, $^{3}$University of Alberta, \\$^{4}$Nanyang Technological University, $^{5}$Dialpad Inc.\\
mohsinulkabir@iut-dhaka.edu, saidulis@yorku.ca, tahmid20@yorku.ca, \\ mnayeem@ualberta.ca, bari0001@e.ntu.edu.sg, enamulh@yorku.ca\\}

\abstract{
Large Language Models (LLMs) have emerged as one of the most important breakthroughs in NLP for their impressive skills in language generation and other language-specific tasks. Though LLMs have been evaluated in various tasks, mostly in English, they have not yet undergone thorough evaluation in under-resourced languages such as Bengali (Bangla). To this end, this paper introduces BenLLM-Eval, which consists of a comprehensive evaluation of LLMs to benchmark their performance in the Bengali language that has modest resources. In this regard, we select various important and diverse Bengali NLP tasks, such as text summarization, question answering, paraphrasing, natural language inference, transliteration, text classification, and sentiment analysis for zero-shot evaluation of popular LLMs, namely, GPT-3.5, LLaMA-2-13b-chat, and Claude-2. Our experimental results demonstrate that while in some Bengali NLP tasks, zero-shot LLMs could achieve performance on par, or even better than current SOTA fine-tuned models; in most tasks, their performance is quite poor (with the performance of open-source LLMs like LLaMA-2-13b-chat being significantly bad) in comparison to the current SOTA results. Therefore, it calls for further efforts to develop a better understanding of LLMs in modest-resourced languages like Bengali. 
 \\ \newline }

\begin{document}
\maketitleabstract

\section{Introduction}

Since the advent of pre-trained language models \cite{devlin-etal-2019-bert, Liu2019RoBERTaAR,rogers-etal-2020-primer}, NLP has witnessed revolutionary advancements over the years. These pre-trained language models have produced SOTA results on a variety of NLP tasks with little task-specific fine-tuning. This leads to the development of various pre-trained language models specialized in the Bengali language, such as BanglaBERT \cite{bhattacharjee-etal-2022-banglabert}, BanglaT5 \cite{bhattacharjee-etal-2023-banglanlg}, and etc. These models have demonstrated exciting progress in many downstream Bengali NLP tasks \cite{ekram-etal-2022-banglarqa, akash-etal-2023-shironaam}. However, one major concern for these pre-trained models is that they require fine-tuning using domain-specific large annotated datasets, which is challenging for Bengali due to its under-representation in the NLP domain \cite{joshi-etal-2020-state, Chakraborty_Nayeem_Ahmad_2021, chowdhury-etal-2021-unsupervised}  despite being the sixth most spoken language in the world with over {300} million native speakers \cite{bengali-language-article}.


Recent developments in large language models (LLMs) \cite{Brown2020LanguageMA, Shoeybi2019MegatronLMTM, Rae2021ScalingLM, Zhang2022OPTOP} have transformed the landscape in NLP. These LLMs, with parameter sizes exceeding a hundred billion, leverage the in-context learning capability to achieve impressive performance in few-shot and zero-shot learning scenarios without the need for task-specific fine-tuning. This capability makes it possible to reduce the need for the annotation of huge datasets, allowing the model to perform well on tasks that it was not trained on. 

Despite the impressive capabilities of LLMs, they may still frequently generate untruthful facts that diverge from the original input \cite{Ouyang2022TrainingLM}. To address this issue, the Reinforcement Learning from Human Feedback (RLHF) mechanism has been utilized to help LLMs generate honest and harmless responses. 
ChatGPT and other recently proposed LLMs like PaLM-2, Claude-2, LLaMA-2-chat \cite{touvron2023llama2, anil2023palm2, jahan2024comprehensive} are trained via leveraging this RLHF technique to mitigate various limitations of the previous generation LLMs and gained widespread popularity. While these LLMs are trained in multiple languages, English possesses the majority of the training data. Though ChatGPT like LLMs has demonstrated strong zero-shot performance in various NLP tasks in English \cite{laskar-etal-2023-systematic,laskar-etal-2023-building} and some other languages \cite{Lai2023ChatGPTBE} and domains \cite{jahan-etal-2023-evaluation,jahan2024comprehensive,fu2024tiny}, these LLMs are yet to be investigated in the widely spoken, yet modest-resourced,
 Bengali language domain. 

In this regard, we aim to assess the efficacy of LLMs in prevalent downstream NLP tasks specific to the Bengali language, a domain that has not been thoroughly examined compared to the numerous LLM evaluations conducted for English and other Indo-European languages. Due to the lack of task-specific annotated datasets in Bengali, we conducted a zero-shot evaluation with LLMs to investigate if LLMs could be utilized to address the scarcity of large annotated datasets in Bengali. Thus, the findings from this paper would give insights into the capabilities and limitations of LLMs in Bengali, which may pave the way to implement LLMs in real-world applications like Bengali Chatbots.
To this end, we present BenLLM-Eval - a comprehensive benchmark that evaluates the zero-shot performance of various LLMs on diverse NLP tasks in the Bengali language, including text summarization, question answering, paraphrasing, natural language inference, transliteration, text classification, and sentiment analysis. The evaluation incorporates carefully crafted prompts to ensure rigorous assessment of the following three popular LLMs: GPT-3.5, LLaMA-2-13b-chat, and Claude-2, and compare them with SOTA fine-tuned models. To the best of our knowledge, this is the first work that evaluates  LLMs in the Bengali language. Our experimental results in this paper are summarized as follows: 


\begin{itemize}
    \item Despite some exceptional cases, the zero-shot performance of LLMs is generally inferior compared to the SOTA fine-tuned models across the majority of the tasks in our evaluation.
    Given the substantial performance disparities observed, it is reasonable to deduce that LLMs, in their current form, are not suitable for serving as a comprehensive solution for modest-resourced tasks in Bengali\footnote{We share the LLM-generated responses, prompts, and parsing scripts for all seven tasks here: \href{https://github.com/saidul-islam98/BenLLMeval}{https://github.com/saidul-islam98/BenLLMeval}}. 
   \item Considering LLMs remarkable performance in zero-shot scenarios within the English language and its subpar performance in modest-resource languages like Bengali, this paper emphasizes the significance of investigating the limitations of LLMs tailored to diverse modest-resource language groups. 
\end{itemize}

\begin{table*}
\centering
\tiny
\begin{tabular}{{p{.17\linewidth}p{.12\linewidth}p{.16\linewidth}p{.44\linewidth}}}
\hline
\textbf{Dataset} & \textbf{Type} & \textbf{Data Split (Train / Valid / Test)} & \textbf{Prompt}\\ \hline

XL-Sum \cite{hasan-etal-2021-xlsum} & Text Summarization & 8102 / 1012 / 1012 & Please provide an one-sentence summary of the following Bangla text input. The input will be a long Bangla paragraph, the output should be a short Bangla paragraph summarizing only the vital information of the input text in one sentence. Please make sure that the output contains the most essential statistical data.
Note: Please do not provide anything other than the summarized Bangla output. \newline \texttt{[INPUT:]}\\
\hline
SQuAD\_Bangla \cite{bhattacharjee-etal-2022-banglabert} & Question-Answering & 118k / 2.5k / 2.5k & Please provide an answer to the input Bangla question based on the given Bangla context. The input will contain a Bangla question followed by a context. The output should be the answer in Bangla based on the context. Note: Please do not provide anything other than the Bangla answer to the question. \newline  \texttt{[CONTEXT:]}  \newline \texttt{[QUESTION:]}\\
\hline
IndicParaphrase \cite{Kumar2022IndicNLGBM} & Paraphrasing & 890k / 10k / 10k & Please provide paraphrasing of the following input Bangla text. The input will be a complex Bangla sentence, the output should be a paraphrased Bangla sentence maintaining the original information of the input text unchanged.
Note: Please do not provide anything except the paraphrased Bangla output. \newline  \texttt{[INPUT:]}\\
\hline
BNLI \cite{bhattacharjee-etal-2022-banglabert} & Natural Language \newline Inference (NLI) & 381k / 2.42k / 4.9k & Please determine the logical relationship between the given hypothesis and premise. The input will consist of two sentences written in the Bangla language. The first sentence represents the premise, while the second sentence represents the hypothesis.  Your task is to determine whether the hypothesis is false (contradiction), true (entailment), or inconclusive (neutral) given the premise. Please output a number indicating the logical relationship between them: 0 for false (contradiction), 1 for true (entailment), and 2 for inconclusive (neutral) for neutrality. 
Note: Please avoid providing any additional information beyond the logical relationship. \newline  \texttt{[PREMISE:]} \newline \texttt{[HYPOTHESIS:]}\\
\hline
Dakshina \citep{roark2020dakshina} & Transliteration \newline (single-word: lexicon) & - / - / 9.2k & Task Description: Please provide the transliteration in native Bengali script for the input word. 
The input will be a word written in Latin script and the output should be the transliterated Bengali word of the given input. Please note that you are not asked to provide translation of the input word, only provide the Bengali transliteration for the given input. \newline
Note: Your response should include only the transliterated word in the native Bengali language. Please do not add any explanation with the output. \newline
\texttt{[INPUT:]} \\
\hline
Dakshina \citep{roark2020dakshina} & Transliteration \newline (full sentence) & 25k / 5k / 5k & Task Description: Please provide the transliteration in native Bengali script for the input sentence. The input will be a sentence written in Latin script and the output should be the transliterated Bengali sentence of the given input. Please do not provide the translation of the input sentence, only provide the Bengali transliteration for the given input. \newline
Note: Your response should include only the transliterated sentence in the native Bengali language. Please do not add any explanation with the output. \newline
\texttt{[INPUT:]} \\
\hline
 Soham News Article Classification \cite{kakwani-etal-2020-indicnlpsuite} & Text Classification & 11284 / 1411 / 1411 &
 
 For the Bengali news article given in the input, identify the appropriate section title for the article from the following classes: kolkata, state, sports, national, entertainment, international. 
 Note: Do not output any unnecessary words other than just the section title. The response should be in English language and should be one word.
\newline  \texttt{[INPUT:]}     
 \\
\hline
IndicSentiment \cite{doddapaneni-IndicXTREME} & Sentiment Analysis & - / 156 / 1000 & For the given Input, is the sentiment in the input positive or negative? 
Note: Please do not output anything other than the sentiment. Exclude any word like, Sentiment in the response. 
\newline \texttt{[INPUT:]}\\
\hline
SentNoB \cite{islam-etal-2021-sentnob-dataset} & Sentiment Analysis & 12575 / 1567 / 1586 & For the given Input, is the sentiment in the input positive or negative or neutral? 
Note: Please do not output anything other than the sentiment. Exclude any word like, Sentiment in the response.
\newline \texttt{[INPUT:]} \\
\hline
\end{tabular}
    
\caption{\small{
Datasets Details with our Prompts for Various Tasks.
}}
\label{tab:dataset details}
\end{table*}

\section{Methodology}
The objective of our study is to assess the efficacy of LLMs in the context of NLP tasks specific to the Bengali language.
We cover {7} diverse and important Bengali NLP tasks: \textbf{(i)} Text Summarization, \textbf{(ii)} Question Answering (QA), \textbf{(iii)} Paraphrasing, \textbf{(iv)} Natural Language Inference (NLI), \textbf{(v)} Transliteration, \textbf{(vi)} Text Classification, and \textbf{(vii)} Sentiment Analysis over {8} benchmark datasets. For this purpose, we evaluate ChatGPT (GPT-3.5\footnote{\href{https://platform.openai.com/docs/models/gpt-3-5}{https://platform.openai.com/docs/models/gpt-3-5}}), Claude-2\footnote{\href{https://www.anthropic.com/index/claude-2}{https://www.anthropic.com/index/claude-2}}, and LLaMA-2-13b-Chat \cite{touvron2023llama2} models. 
Similar to the prior work on LLM evaluation \cite{laskar-etal-2023-systematic,jahan-etal-2023-evaluation}, we also focus on designing a zero-shot learning setting to benchmark the performance of the LLMs. 

We prepare a task instruction $T$ for a given test sample $X$ and concatenate the text in the test sample with the task instruction to construct the prompt $P$. The prompt $P$ is then passed as input to the LLMs, which generates the response $R$. A comprehensive description of the tasks, datasets, and prompts devised for evaluating each specific task is presented below and also summarized in Table \ref{tab:dataset details}. Next, we briefly discuss each task. 



    \textbf{Text Summarization:} Summarization is the process of automatically generating a concise and coherent summary of a longer text document \cite{nayeem-chali-2017-extract, 10.1145/3132847.3133106,laskar2022domain}, preserving the most important information while reducing the length \cite{nayeem-etal-2018-abstractive}. 
    In this paper, we evaluate the XL-Sum dataset \cite{hasan-etal-2021-xlsum} that consists of 1 million manually annotated data samples from {44} languages. We only took the Bengali samples for evaluation.

    \textbf{Question Answering:} For the question-answering task, we evaluate the performance of LLMs on the SQuAD\_Bangla dataset \cite{bhattacharjee-etal-2022-banglabert}. This dataset was constituted using two benchmark English datasets: SQuAD 2.0 \cite{williams-etal-2018-squad2} and TyDi QA \cite{clark-etal-2020-tydi}. The objective of this task is to determine whether the answer to a given question $Q$ can be inferred from the reference context $C$. We provide the reference context along with the question and ask LLMs whether they can infer the answer to the question from the given reference. 

    \textbf{Paraphrasing:} The paraphrasing task aims to generate a paraphrase of the input text. To evaluate this task, we choose the Bengali samples from the IndicParaphrase dataset \cite{Kumar2022IndicNLGBM}, which is the largest Indic paraphrasing dataset across {11} Indic languages \cite{indic-languages-article}. 

    \textbf{Natural Language Inference:} Natural Language Inference (NLI) aims to predict the entailment/contradiction relations between two input sentences: a premise and a hypothesis. To evaluate LLMs for Bengali NLI, we utilize the BNLI dataset \cite{bhattacharjee-etal-2022-banglabert} (curated from the benchmark XNLI dataset \cite{conneau-etal-2018-xnli}) that provides annotated data with three categories: \textit{Entailment, Contradiction,} and \textit{Neutral}.

    \textbf{Transliteration:} Transliteration is the process of converting texts from one script to another based on phonetic similarity. We utilize the \emph{Dakshina dataset} \citep{roark2020dakshina} to evaluate transliteration errors. 
    Errors can be \textbf{(i)} \emph{Substitutions}, \textbf{(ii)} \emph{Insertions}, or \textbf{(iii)} \emph{Deletions}. A substitution error occurs when the predicted transliteration differs from the reference transliteration. An insertion error happens when extra characters/words are present in the predicted transliteration. A deletion error occurs when characters/words are missing from the predicted transliteration. The error rate is calculated as:
    \begin{align*}
    \small
    \text{error rate} = \frac{\text{substitutions + deletions + insertions}}{\text{total reference tokens}} * 100
    \label{form:error_rate}
    \end{align*}
    Two types of error rates are proposed for this task in \emph{Dakshina dataset} \citep{roark2020dakshina}: the Character-Error Rate (\textbf{CER}) treats each token as an individual Unicode character, while the Word-Error Rate (\textbf{WER}) treats each token as a substring separated by whitespace. We report both CER and WER for single-word transliteration, and WER for full-sentence transliteration.

    \textbf{Text Classification:} This task refers to the classification of the category for a given input text. To evaluate the text classification capability of LLMs, we use the \textit{Soham Bengali News Classification} dataset that is included in the IndicGLUE \cite{kakwani-etal-2020-indicnlpsuite} benchmark. This dataset contains six news categories i.e., \textit{kolkata, state, national, international, sports,} and \textit{entertainment}. 
    
    \textbf{Sentiment Analysis:} We evaluate the Sentiment Analysis capability of the LLMs with two datasets: SentNoB \cite{islam-etal-2021-sentnob-dataset}, and IndicSentiment \cite{doddapaneni-IndicXTREME}. The SentNoB dataset comprises texts that are informally written in Bengali, collected from public comments on news and videos in social media covering a wide range of domains like, \textit{politics}, \textit{agriculture}, \textit{education} etc. The second dataset, IndicSentiment, was manually curated for the IndicXTREME benchmark and contains product reviews of multiple categories. We only used the Bengali samples from the dataset out of the {12} Indic languages.

\begin{table*}[t]
\centering
\resizebox{16cm}{!} 
{ 
\begin{tabular}{lccclclclclclclccc}
\hline
 &
  \multicolumn{3}{c}{\textbf{XL-Sum ({TS})}} &
   &
  \multicolumn{1}{l}{\textbf{SQuAD\_Bangla ({QA})}} &
   &
  \textbf{IndicPara ({PP})} &
   &
  \textbf{BNLI ({NLI})} & &
  \textbf{SNAC ({TC})}
   &
   &
  \textbf{IndicSent ({SA})} &
  & \multicolumn{3}{c}{\textbf{SentNoB ({SA})}}\\ \cline{2-4}
   \cline{6-6} \cline{8-8} \cline{10-10} \cline{12-12} \cline{14-14} \cline{16-18} 
Model                  & R-1            & R-2            & R-L            &  &  EM / F1                &  & BLEU           &  & Acc.      &      & Acc.  &  & Acc. & & P &  R & F1 \\ \hline
\textbf{GPT-3.5}       & 20.19          & 5.81           & 15.53          &  & 44.85/78.67          &  & 2.81           &  & 52.71         &     & 48.47 &  & \textbf{90.20 }                &  & \textbf{57.70}  & 54.56 & 53.17 \\ 
\textbf{LLaMA-2-13b-chat} & 0.41          & 0.14           & 0.34          &  & 31.73/67.95          &  & 0.01           &  & 42.37         &   & 29.27 &  & 69.16                &  & 48.39  & 48.49 & 48.43 \\ 
\textbf{Claude-2} & 20.79          & 5.55           & 16.47          &  & 49.92/79.04         &  & 1.89           &  & 32.20         &  & 48.61   &  & 88.48                &  & 53.28  & 54.38 & 52.79 \\  \hline
\textbf{mT5} \cite{hasan-etal-2021-xlsum}           & \textbf{28.32} & \textbf{11.43} & \textbf{24.23} &  &  \multicolumn{1}{l}{} &  & 4.45           &  &       -        &  & - &  & - &  & - & - & -\\ 
\textbf{BanglaBERT} \cite{bhattacharjee-etal-2022-banglabert}    &        -        &        -        &        -        &  & 72.63/\textbf{79.34  }        &  &       -        &  & \textbf{82.8} &  & - &  & - &  & - & - & -\\
\textbf{BanglishBERT} \cite{bhattacharjee-etal-2022-banglabert}  &        -        &          -      &         -       &  & 72.43/78.40          &  &       -         &  & 80.95         &  & -  &  & - &  & - & - & -\\
\textbf{XLM-R (Large)} \cite{bhattacharjee-etal-2022-banglabert} &          -      &         -       &         -       &  & \textbf{73.15}/79.06 &  &         -       &  & 82.4      &    & - &  & - &  & - & - & -\\
\textbf{XLM-R} \cite{kakwani-etal-2020-indicnlpsuite, doddapaneni-IndicXTREME} &         -       &          -      &            -    &  &  &  &            -    &  &      -     &  & \textbf{87.60} &   & 85.8 &  & - & - & -\\
\textbf{IndicBART} \cite{Kumar2022IndicNLGBM}    &      -          &         -       &       -         &  & \multicolumn{1}{l}{} &  & \textbf{11.57} &  &       -        &  & - &  & - &  & - & - & - \\
\textbf{IndicBERT} \cite{kakwani-etal-2020-indicnlpsuite, doddapaneni-IndicXTREME} &         -       &         -       &         -       &  &  &  &          -      &  &     -     &  & 78.45 &  & 89.3 &  & - & - & -\\
\textbf{mBERT} \cite{kakwani-etal-2020-indicnlpsuite, doddapaneni-IndicXTREME} &        -        &        -        &        -        &  &  &  &             -   &  &    -      &  & 80.23 &  & 72.0 &  & 49.58 & 56.43 & 52.79\\
\textbf{Bi-LSTM + Attn. (\textit{w/} FastText)} \cite{islam-etal-2021-sentnob-dataset} &         -       &        -       &         -       &  &  &  &       -         &  &      -    &  & - &   & - &  & 52.24 & 63.09 & 57.15\\
\textbf{Bi-LSTM + Attn. (\textit{w/} Rand init)} \cite{islam-etal-2021-sentnob-dataset} &        -        &          -      &       -         &  &  &  &         -       &  &     -     &  & -  &   & - &  & 56.16 & \textbf{64.97} & \textbf{60.25} \\
\hline
\end{tabular}%
}
\caption{\small{Performance Comparison between zero-shot LLMs \& fine-tuned SOTA models on Text Summarization  ({TS}), Question Answering  ({QA}), Paraphrasing ({PP}), Natural Language Inference ({NLI}), Text Classification ({TC}), and Sentiment Analysis ({SA}). EM, Acc., P, R, and F1 denote Exact Match, Accuracy, Precision, Recall, and F1 score respectively. Best results are \textbf{boldfaced}.}}
\label{tab: performance}
\end{table*}
\begin{table*}[t]
\centering
\resizebox{16cm}{!} 
{ 
\begin{tabular}{lcclcclcclclcclcclcc}
\hline
\textbf{Task} &
  \multicolumn{2}{c}{\textbf{Pair 6-gram}} &
   &
  \multicolumn{2}{c}{\textbf{LSTM}} &
   &
  \multicolumn{2}{c}{\textbf{Transformer}} &
   &
  \textbf{Noisy Channel} &
   &
  \multicolumn{2}{c}{\textbf{GPT-3.5}} &
   &
  \multicolumn{2}{c}{\textbf{LLaMA-2-13b}} &
   &
  \multicolumn{2}{c}{\textbf{Claude 2}} \\ \cline{2-3} \cline{5-6} \cline{8-9} \cline{11-11} \cline{13-14} \cline{16-17} \cline{19-20} 
\multicolumn{1}{c}{\textbf{}} &
  \textbf{CER (↓)} &
  \textbf{WER (↓)} &
   &
  \textbf{CER (↓)} &
  \textbf{WER (↓)} &
   &
  \textbf{CER (↓)} &
  \textbf{WER (↓)} &
   &
  \textbf{WER (↓)} &
   &
  \textbf{CER (↓)} &
  \textbf{WER (↓)} &
   &
  \textbf{CER (↓)} &
  \textbf{WER (↓)} &
   &
  \textbf{CER (↓)} &
  \textbf{WER (↓)} \\ \hline
\textbf{\begin{tabular}[c]{@{}l@{}}Lexicon\\\end{tabular}} &
  14.2 &
  54.0 &
   &
  13.9 &
  54.7 &
   &
  \textbf{13.2} &
  \textbf{50.6} &
   &
  - &
   &
  \textbf{18.1} &
  \textbf{60.6} &
   &
  39.85 &
  80.72 &
   &
  23.16 &
  68.07 \\ \hline
\textbf{\begin{tabular}[c]{@{}l@{}}Sentence\\\end{tabular}} &
  - &
  39.7 &
   &
  - &
  - &
   &
  - &
  37.6 &
   &
  \textbf{25.8} &
   &
  - &
  \textbf{29.9} &
   &
  - &
  66.54 &
   &
  - &
  38.10 \\ \hline
\end{tabular}%
}
\caption{\small{
Single-word and Full-sentence Transliteration results. Here, the baseline results are adopted from \citet{roark2020dakshina}. 
Best results are \textbf{boldfaced} and lower (↓) is better.}}
\label{tab:translit_eval}
\end{table*}

\section{Results and Discussion}
We report the performance of LLMs for different tasks and compare their performance with the current SOTA fine-tuned models (see Table \ref{tab: performance}). 






\textbf{Text Summarization Evaluation:} In the XL-Sum dataset, we find that none of the LLMs could outperform the current SOTA mT5 model. Among the LLMs, Claude-2 performs the best in terms of ROUGE-1 and ROUGE-L, while GPT-3.5 performs the best in terms of ROUGE-2. We also find that GPT-3.5 tends to generate much longer summaries (229 words on average) than the gold reference summaries (148 words on average), while the length of the Claude-2 generated summaries (137 words on average) is more consistent with the gold reference summaries (148 words on average).  
Moreover, to find out why LLaMA-2-13b-chat achieves very low ROUGE scores, we manually reviewed the responses 
and found out that it ended up generating all the summaries in English. By translating the summaries to Bengali using Google's Translation API \cite{google-translate-api},
the R-1, 2, and L scores are increased to  4.69, 0.61, and 3.61, respectively, for LLaMA-2-13b-chat.         

\textbf{Question Answering Evaluation:} Since LLMs frequently generate responses that may not be an exact match of the gold label but are nonetheless correct, our QA evaluation utilizes human intervention to compute the F1 score. We find that Claude-2 and GPT-3.5 achieve performance almost similar to the current SOTA results in terms of F1. However, the performance of all LLMs is quite poor in terms of Exact Match due to generating descriptive responses, as well as paraphrases. 

\textbf{Paraphrasing Evaluation:} For the paraphrasing task, we 
observe a very low BLEU score for all LLMs, which is a phenomenon similar to what happened with the \textbf{EM} metric for the QA task. Note that the BLEU score also computes word-level similarity, while such limitations of word-based similarity metrics have also been noticed LLMs are evaluated in English datasets \cite{laskar-etal-2023-systematic}. 

\begin{figure*}[t]
\centering
\begin{subfigure}{.32\textwidth}
    \centering
    \includegraphics[scale=0.4]{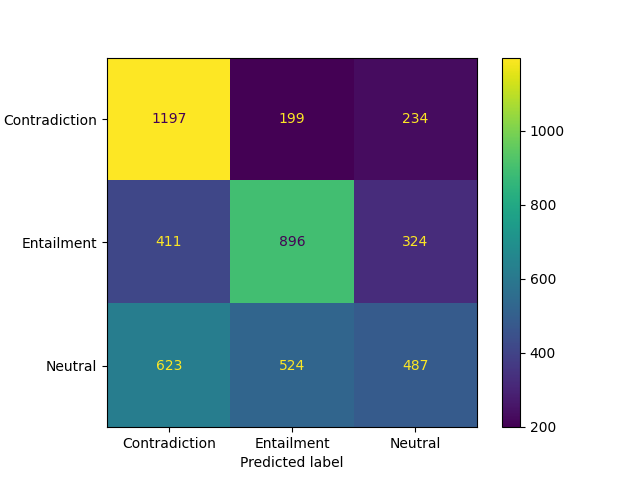}  
    \caption{GPT-3.5}
    \label{SUBFIGURE LABEL 1}
\end{subfigure}
\begin{subfigure}{.32\textwidth}
    \centering
    \includegraphics[scale=0.4]{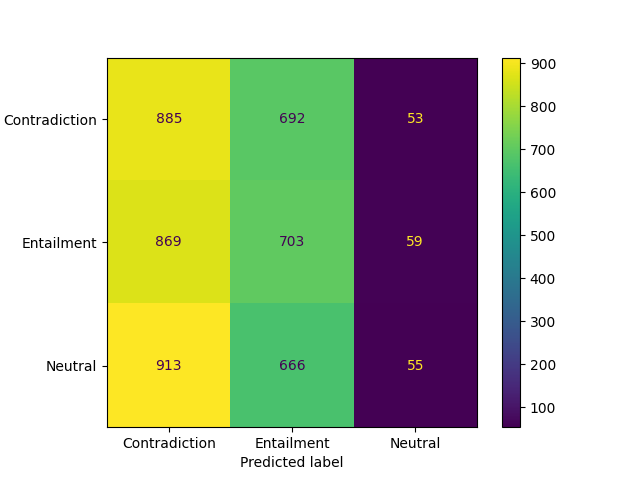}  
    \caption{Llama-2-13b-chat}
    \label{SUBFIGURE LABEL 2}
\end{subfigure}
\begin{subfigure}{.32\textwidth}
    \centering
    \includegraphics[scale=0.4]{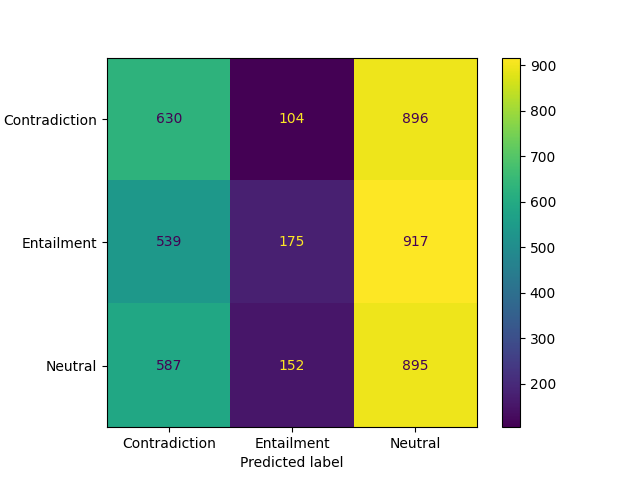}  
    \caption{Claude-2}
    \label{SUBFIGURE LABEL 3}
\end{subfigure}
\caption{Confusion matrices for different LLMs on the BNLI dataset for the NLI task. }
\label{FIGURE LABEL}
\end{figure*}

\textbf{Natural Language Inference Evaluation:} We observe that while GPT-3.5 achieves the best results among all LLMs on the BNLI dataset, it is still much lower in comparison to the current SOTA BanglaBERT. Surprisingly, we find that Claude-2 is the worst performer in this task while being outperformed by the much smaller LLaMA-2-13b-chat model. To further investigate in which cases LLMs perform poorly in the NLI task, we demonstrate their predictions using the Confusion Matrix in Figure \ref{FIGURE LABEL} and find that LLaMA-2-13b-chat was quite bad while predicting the \textit{Neutral} types and Claude-2 performed poorly while predicting the \textit{Entailment} types. In the case of GPT-3.5, it performed poorer while predicting the \textit{Neutral} type class in comparison to the other types.  

\textbf{Transliteration Evaluation:} 
From Table \ref{tab:translit_eval}, we find that GPT-3.5 achieves superior performance over other LLMs in single-word transliteration. While it fails to outperform the existing results, it still achieves competitive performance compared to the prior SOTA models. This is an interesting observation given the intricate structure and infrequent usage of lexicons in the test set.
In full-sentence transliteration, 
GPT-3.5 makes fewer word errors and character errors (60.6\% \& 18.1\%) compared to other LLMs ((80.72\% \& 39.85\%) for Llama-2-13b-chat and (68.07\% \& 23.16\%) for Claude-2). However, the results are quite poor compared to other SOTA models.

\textbf{Text Classification Evaluation:} LLMs performed poorly on the Soham News Article classification dataset, with the best-performing LLM in this dataset, the Claude-2 model achieving only {48.61}\% accuracy, followed by GPT-3.5 achieving 48.47\% accuracy while the LLaMA-2-13b-chat being the worst performer with an accuracy of {29.27}\%.  On the contrary, the XLM-R model, which is the SOTA for this task, obtained an accuracy of {87.60}\% in this dataset. While we achieved these results by providing the class labels alongside our prompts (see Table \ref{tab:dataset details}), if the class labels were not mentioned, the accuracy dropped to 20.76\%, 18.36\%, and 1.06\% for Claude-2, GPT-3.5, and LLaMA-2-13b-chat, respectively. This indicates the importance of using descriptive prompts to improve the zero-shot performance of LLMs.  

\textbf{Sentiment Analysis Evaluation:} In the sentiment classification task, we observe that GPT-3.5 performed exceptionally well on the IndicSentiment dataset \cite{doddapaneni-IndicXTREME}, attaining a new SOTA accuracy of {90.20}\% 
While we also find that the Claude-2 model achieved impressive performance ({88.48}\% accuracy), the LLaMA-2-13b-chat model performed much poorer. We also find a similar trend for sentiment analysis in the SentNoB \cite{islam-etal-2021-sentnob-dataset} dataset. In this dataset, to ensure a fair comparison with the current SOTA, we use the precision, recall, and F1 scores as evaluation metrics. While none of the LLMs could outperform the current SOTA, all of them (especially GPT-3.5 and Claude-2) achieved performance comparable to the SOTA results. This is consistent with LLMs' performance for this task in English \cite{laskar-etal-2023-systematic}.  


\section{Task Contamination Analysis}
Task contamination analysis \cite{li2023task} is essential to ensure a fair model evaluation since it helps identify a model's prior exposure to test tasks on its training data. 
Inspired by the work of \citet{li2023task}, we include task contamination analysis in our evaluation to appropriately assess the performance of the LLMs. We utilize two methods: \textit{Task Example Extraction (TEE)} and \textit{Membership Inference (for generative tasks like summarization and paraphrasing)} to verify the evidence of task contamination. \textbf{\textit{TEE}} involves retrieving task examples from instruction-tuned models, 
although it doesn't require an exact match with existing training data, helps identify potential task contamination in zero and few-shot learning scenarios, as any task examples found indicate possible prior exposure. Additionally, \textbf{\textit{Membership Inference}} is a 
method used for generation tasks, where the model's generated content is checked against the original dataset for exact matches. If a match is found, it implies that the content was part of the LLM's training data, indicating direct exposure rather than general learning ability.

At first, we applied TEE to all tasks without explicitly mentioning the dataset name. Our findings reveal that only GPT-3.5 could generate examples related to these tasks (except Natural Language Inference), while Claude-2 and LLaMA-2-13b-chat models failed to extract task examples for any tasks. Therefore, there is a possibility that such tasks were already included in the pre-training data of GPT-3.5. Regarding the BNLI dataset where no models could extract any task examples, we find that the premises, hypotheses, and labels generated by all LLMs for Bengali were significantly inaccurate, providing evidence that contamination did not occur. In terms of extracting task examples in the transliteration task, we find that among LLMs, only GPT-3.5 could extract the task examples for both word-level and sentence-level transliteration. 


Finally, we applied the Membership Inference technique in the generative tasks (e.g., summarization and paraphrase generation) to analyze task contamination. In the summarization task, none of the LLMs produced an output that aligned with the test data labels. However, on the paraphrasing task, GPT-3.5 produced around 50 exact match instances, while Claude-2 produced 30 and LLaMA-2-13b-chat produced 15 exact matches of the generated outputs and test labels. So it is possible that the test set of the IndicPara dataset was exposed to the LLMs while training. 

In summary, contamination could be an issue with the GPT-3.5 model in Sentiment Analysis, Text Classification, Summarization, and QA tasks, while all the models, i.e., GPT-3.5, LLaMA-2-13b-chat, and Claude-2 were affected by task contamination in the Paraphrasing task. However, in Natural Language Inference, we did not see any evidence of task contamination.

\section{Conclusions and Future Work}
In this paper, we introduce BenLLM-Eval, which provides a comprehensive zero-shot evaluation of LLMs 
on seven benchmark NLP tasks to understand the capability of LLMs in the modest-resourced Bengali language. The results revealed that while in some tasks, zero-shot closed-source LLMs like GPT-3.5 or Claude-2 perform on par (e.g., summarization) or even outperform (e.g., sentiment analysis) current SOTA models, in most tasks the LLMs exhibited considerably lower performance compared to supervised fine-tuned models. We also observed that the open-source LLaMA-2-13b-chat model performed significantly poorer in most tasks. Thus, open-source LLMs should be extensively evaluated on low to modest-resource languages to ensure a proper understanding of their capabilities and limitations. 
Our findings also reveal that in order to achieve optimal performance across languages, LLMs should also be trained on corpus covering 
various languages along with English. In the future, we intend to expand our experiments by including additional low to modest-resource languages, tasks, datasets, and settings to obtain broader insights. 

\section*{Limitations} The datasets used to train LLMs like GPT-3.5 or Claude-2 are not disclosed; consequently, some datasets used to evaluate the model may or may not have existed during model training. In addition, the API used to evaluate the GPT-3.5 model is based on OpenAI's \textit{GPT-3.5 turbo}. Although an improved version GPT-4 is also available, GPT-4 is significantly more expensive than GPT-3.5. Thus, we could not use GPT-4 in this work. 
Other than these limitations, this study will provide a practical direction for future research into the performance of LLMs in modest-resourced languages in the NLP domain.

\section*{Ethics Statement} This study assesses the performance of LLMs in seven benchmark NLP tasks in the Bengali language wherein LLMs are tasked with generating an output based on the information given in the input. Therefore, no prompt or information was presented to LLMs that could result in the generation of any output raising any ethical issues or unwanted biases. There will be no licensing issue as all the datasets that are utilized in this research are publicly available.

\section*{Acknowledgements} This research is supported by the Generic (Minor/Startup/Other) research funds of York University. We also thank Anthropic for providing free access to the Claude-2 model, the Canada Foundation for Innovation grant and Compute Canada for its computing resources. Mir Tafseer Nayeem is also supported by the Huawei PhD Fellowship.


\nocite{Ando2005,augenstein-etal-2016-stance,andrew2007scalable,rasooli-tetrault-2015,goodman-etal-2016-noise,harper-2014-learning}

\bibliographystyle{lrec-coling2024-natbib}
\bibliography{lrec-coling2024-example}

\begin{thebibliography}{41}
\expandafter\ifx\csname natexlab\endcsname\relax\def\natexlab#1{#1}\fi

\bibitem[{Akash et~al.(2023)Akash, Nayeem, Shohan, and
  Islam}]{akash-etal-2023-shironaam}
Abu~Ubaida Akash, Mir~Tafseer Nayeem, Faisal~Tareque Shohan, and Tanvir Islam.
  2023.
\newblock \href {https://aclanthology.org/2023.eacl-main.4} {Shironaam:
  {B}engali news headline generation using auxiliary information}.
\newblock In \emph{Proceedings of the 17th Conference of the European Chapter
  of the Association for Computational Linguistics}, pages 52--67, Dubrovnik,
  Croatia. Association for Computational Linguistics.

\bibitem[{Anil et~al.(2023)Anil, Dai, Firat, Johnson, Lepikhin, Passos,
  Shakeri, Taropa, Bailey, Chen et~al.}]{anil2023palm2}
Rohan Anil, Andrew~M Dai, Orhan Firat, Melvin Johnson, Dmitry Lepikhin,
  Alexandre Passos, Siamak Shakeri, Emanuel Taropa, Paige Bailey, Zhifeng Chen,
  et~al. 2023.
\newblock Palm 2 technical report.
\newblock \emph{arXiv preprint arXiv:2305.10403}.

\bibitem[{Bhattacharjee et~al.(2022)Bhattacharjee, Hasan, Ahmad, Mubasshir,
  Islam, Iqbal, Rahman, and Shahriyar}]{bhattacharjee-etal-2022-banglabert}
Abhik Bhattacharjee, Tahmid Hasan, Wasi Ahmad, Kazi~Samin Mubasshir, Md~Saiful
  Islam, Anindya Iqbal, M.~Sohel Rahman, and Rifat Shahriyar. 2022.
\newblock \href {https://doi.org/10.18653/v1/2022.findings-naacl.98}
  {{B}angla{BERT}: Language model pretraining and benchmarks for low-resource
  language understanding evaluation in {B}angla}.
\newblock In \emph{Findings of the Association for Computational Linguistics:
  NAACL 2022}, pages 1318--1327, Seattle, United States. Association for
  Computational Linguistics.

\bibitem[{Bhattacharjee et~al.(2023)Bhattacharjee, Hasan, Ahmad, and
  Shahriyar}]{bhattacharjee-etal-2023-banglanlg}
Abhik Bhattacharjee, Tahmid Hasan, Wasi~Uddin Ahmad, and Rifat Shahriyar. 2023.
\newblock \href {https://aclanthology.org/2023.findings-eacl.54} {{B}angla{NLG}
  and {B}angla{T}5: Benchmarks and resources for evaluating low-resource
  natural language generation in {B}angla}.
\newblock In \emph{Findings of the Association for Computational Linguistics:
  EACL 2023}, pages 726--735, Dubrovnik, Croatia. Association for Computational
  Linguistics.

\bibitem[{Brown et~al.(2020)Brown, Mann, Ryder, Subbiah, Kaplan, Dhariwal,
  Neelakantan, Shyam, Sastry, Askell, Agarwal, Herbert-Voss, Krueger, Henighan,
  Child, Ramesh, Ziegler, Wu, Winter, Hesse, Chen, Sigler, Litwin, Gray, Chess,
  Clark, Berner, McCandlish, Radford, Sutskever, and
  Amodei}]{Brown2020LanguageMA}
Tom Brown, Benjamin Mann, Nick Ryder, Melanie Subbiah, Jared~D Kaplan, Prafulla
  Dhariwal, Arvind Neelakantan, Pranav Shyam, Girish Sastry, Amanda Askell,
  Sandhini Agarwal, Ariel Herbert-Voss, Gretchen Krueger, Tom Henighan, Rewon
  Child, Aditya Ramesh, Daniel Ziegler, Jeffrey Wu, Clemens Winter, Chris
  Hesse, Mark Chen, Eric Sigler, Mateusz Litwin, Scott Gray, Benjamin Chess,
  Jack Clark, Christopher Berner, Sam McCandlish, Alec Radford, Ilya Sutskever,
  and Dario Amodei. 2020.
\newblock \href
  {https://proceedings.neurips.cc/paper_files/paper/2020/file/1457c0d6bfcb4967418bfb8ac142f64a-Paper.pdf}
  {Language models are few-shot learners}.
\newblock In \emph{Advances in Neural Information Processing Systems},
  volume~33, pages 1877--1901. Curran Associates, Inc.

\bibitem[{Chakraborty et~al.(2021)Chakraborty, Nayeem, and
  Ahmad}]{Chakraborty_Nayeem_Ahmad_2021}
Susmoy Chakraborty, Mir~Tafseer Nayeem, and Wasi~Uddin Ahmad. 2021.
\newblock \href {https://doi.org/10.1609/aaai.v35i14.17495} {Simple or complex?
  learning to predict readability of bengali texts}.
\newblock \emph{Proceedings of the AAAI Conference on Artificial Intelligence},
  35(14):12621--12629.

\bibitem[{Chowdhury et~al.(2021)Chowdhury, Nayeem, Mim, Chowdhury, and
  Jannat}]{chowdhury-etal-2021-unsupervised}
Radia~Rayan Chowdhury, Mir~Tafseer Nayeem, Tahsin~Tasnim Mim, Md. Saifur~Rahman
  Chowdhury, and Taufiqul Jannat. 2021.
\newblock \href {https://doi.org/10.18653/v1/2021.eacl-main.224} {Unsupervised
  abstractive summarization of {B}engali text documents}.
\newblock In \emph{Proceedings of the 16th Conference of the European Chapter
  of the Association for Computational Linguistics: Main Volume}, pages
  2612--2619, Online. Association for Computational Linguistics.

\bibitem[{Clark et~al.(2020)Clark, Choi, Collins, Garrette, Kwiatkowski,
  Nikolaev, and Palomaki}]{clark-etal-2020-tydi}
Jonathan~H. Clark, Eunsol Choi, Michael Collins, Dan Garrette, Tom Kwiatkowski,
  Vitaly Nikolaev, and Jennimaria Palomaki. 2020.
\newblock \href {https://doi.org/10.1162/tacl_a_00317} {{T}y{D}i {QA}: A
  benchmark for information-seeking question answering in typologically diverse
  languages}.
\newblock \emph{Transactions of the Association for Computational Linguistics},
  8:454--470.

\bibitem[{Conneau et~al.(2018)Conneau, Rinott, Lample, Williams, Bowman,
  Schwenk, and Stoyanov}]{conneau-etal-2018-xnli}
Alexis Conneau, Ruty Rinott, Guillaume Lample, Adina Williams, Samuel Bowman,
  Holger Schwenk, and Veselin Stoyanov. 2018.
\newblock \href {https://doi.org/10.18653/v1/D18-1269} {{XNLI}: Evaluating
  cross-lingual sentence representations}.
\newblock In \emph{Proceedings of the 2018 Conference on Empirical Methods in
  Natural Language Processing}, pages 2475--2485, Brussels, Belgium.
  Association for Computational Linguistics.

\bibitem[{Devlin et~al.(2019)Devlin, Chang, Lee, and
  Toutanova}]{devlin-etal-2019-bert}
Jacob Devlin, Ming-Wei Chang, Kenton Lee, and Kristina Toutanova. 2019.
\newblock \href {https://doi.org/10.18653/v1/N19-1423} {{BERT}: Pre-training of
  deep bidirectional transformers for language understanding}.
\newblock In \emph{Proceedings of the 2019 Conference of the North {A}merican
  Chapter of the Association for Computational Linguistics: Human Language
  Technologies, Volume 1 (Long and Short Papers)}, pages 4171--4186,
  Minneapolis, Minnesota. Association for Computational Linguistics.

\bibitem[{Doddapaneni et~al.(2022)Doddapaneni, Aralikatte, Ramesh, Goyal,
  Khapra, Kunchukuttan, and Kumar}]{doddapaneni-IndicXTREME}
Sumanth Doddapaneni, Rahul Aralikatte, Gowtham Ramesh, Shreya Goyal, Mitesh~M.
  Khapra, Anoop Kunchukuttan, and Pratyush Kumar. 2022.
\newblock \href {http://arxiv.org/abs/2212.05409} {Indicxtreme: A multi-task
  benchmark for evaluating indic languages}.

\bibitem[{Ekram et~al.(2022)Ekram, Rahman, Altaf, Islam, Rahman, Rahman,
  Hossain, and Kamal}]{ekram-etal-2022-banglarqa}
Syed Mohammed~Sartaj Ekram, Adham~Arik Rahman, Md.~Sajid Altaf, Mohammed~Saidul
  Islam, Mehrab~Mustafy Rahman, Md~Mezbaur Rahman, Md~Azam Hossain, and Abu
  Raihan~Mostofa Kamal. 2022.
\newblock \href {https://doi.org/10.18653/v1/2022.findings-emnlp.186}
  {{B}angla{RQA}: A benchmark dataset for under-resourced {B}angla language
  reading comprehension-based question answering with diverse question-answer
  types}.
\newblock In \emph{Findings of the Association for Computational Linguistics:
  EMNLP 2022}, pages 2518--2532, Abu Dhabi, United Arab Emirates. Association
  for Computational Linguistics.

\bibitem[{Fu et~al.(2024)Fu, Laskar, Khasanova, Chen, and TN}]{fu2024tiny}
Xue-Yong Fu, Md~Tahmid~Rahman Laskar, Elena Khasanova, Cheng Chen, and
  Shashi~Bhushan TN. 2024.
\newblock Tiny titans: Can smaller large language models punch above their
  weight in the real world for meeting summarization?
\newblock \emph{arXiv preprint arXiv:2402.00841}.

\bibitem[{Google(2020)}]{google-translate-api}
Google. 2020.
\newblock \href {https://pypi.org/project/googletrans} {Googletrans: Free and
  unlimited google translate api for python}.
\newblock Accessed: March 12, 2024.

\bibitem[{Hasan et~al.(2021)Hasan, Bhattacharjee, Islam, Mubasshir, Li, Kang,
  Rahman, and Shahriyar}]{hasan-etal-2021-xlsum}
Tahmid Hasan, Abhik Bhattacharjee, Md.~Saiful Islam, Kazi Mubasshir, Yuan-Fang
  Li, Yong-Bin Kang, M.~Sohel Rahman, and Rifat Shahriyar. 2021.
\newblock \href {https://doi.org/10.18653/v1/2021.findings-acl.413} {{XL}-sum:
  Large-scale multilingual abstractive summarization for 44 languages}.
\newblock In \emph{Findings of the Association for Computational Linguistics:
  ACL-IJCNLP 2021}, pages 4693--4703, Online. Association for Computational
  Linguistics.

\bibitem[{Islam et~al.(2021)Islam, Kar, Islam, and
  Amin}]{islam-etal-2021-sentnob-dataset}
Khondoker~Ittehadul Islam, Sudipta Kar, Md~Saiful Islam, and Mohammad~Ruhul
  Amin. 2021.
\newblock \href {https://doi.org/10.18653/v1/2021.findings-emnlp.278}
  {{S}ent{N}o{B}: A dataset for analysing sentiment on noisy {B}angla texts}.
\newblock In \emph{Findings of the Association for Computational Linguistics:
  EMNLP 2021}, pages 3265--3271, Punta Cana, Dominican Republic. Association
  for Computational Linguistics.

\bibitem[{Jahan et~al.(2023)Jahan, Laskar, Peng, and
  Huang}]{jahan-etal-2023-evaluation}
Israt Jahan, Md~Tahmid~Rahman Laskar, Chun Peng, and Jimmy Huang. 2023.
\newblock \href {https://doi.org/10.18653/v1/2023.bionlp-1.30} {Evaluation of
  {C}hat{GPT} on biomedical tasks: A zero-shot comparison with fine-tuned
  generative transformers}.
\newblock In \emph{The 22nd Workshop on Biomedical Natural Language Processing
  and BioNLP Shared Tasks}, pages 326--336, Toronto, Canada. Association for
  Computational Linguistics.

\bibitem[{Jahan et~al.(2024)Jahan, Laskar, Peng, and
  Huang}]{jahan2024comprehensive}
Israt Jahan, Md~Tahmid~Rahman Laskar, Chun Peng, and Jimmy~Xiangji Huang. 2024.
\newblock A comprehensive evaluation of large language models on benchmark
  biomedical text processing tasks.
\newblock \emph{Computers in Biology and Medicine}, page 108189.

\bibitem[{Joshi et~al.(2020)Joshi, Santy, Budhiraja, Bali, and
  Choudhury}]{joshi-etal-2020-state}
Pratik Joshi, Sebastin Santy, Amar Budhiraja, Kalika Bali, and Monojit
  Choudhury. 2020.
\newblock \href {https://doi.org/10.18653/v1/2020.acl-main.560} {The state and
  fate of linguistic diversity and inclusion in the {NLP} world}.
\newblock In \emph{Proceedings of the 58th Annual Meeting of the Association
  for Computational Linguistics}, pages 6282--6293, Online. Association for
  Computational Linguistics.

\bibitem[{Kakwani et~al.(2020)Kakwani, Kunchukuttan, Golla, N.C.,
  Bhattacharyya, Khapra, and Kumar}]{kakwani-etal-2020-indicnlpsuite}
Divyanshu Kakwani, Anoop Kunchukuttan, Satish Golla, Gokul N.C., Avik
  Bhattacharyya, Mitesh~M. Khapra, and Pratyush Kumar. 2020.
\newblock \href {https://doi.org/10.18653/v1/2020.findings-emnlp.445}
  {{I}ndic{NLPS}uite: Monolingual corpora, evaluation benchmarks and
  pre-trained multilingual language models for {I}ndian languages}.
\newblock In \emph{Findings of the Association for Computational Linguistics:
  EMNLP 2020}, pages 4948--4961, Online. Association for Computational
  Linguistics.

\bibitem[{Kumar et~al.(2022)Kumar, Shrotriya, Sahu, Dabre, Puduppully,
  Kunchukuttan, Mishra, Khapra, and Kumar}]{Kumar2022IndicNLGBM}
Aman Kumar, Himani Shrotriya, Prachi~Rani Sahu, Raj Dabre, Ratish Puduppully,
  Anoop Kunchukuttan, Amogh Mishra, Mitesh~M. Khapra, and Pratyush Kumar. 2022.
\newblock Indicnlg benchmark: Multilingual datasets for diverse nlg tasks in
  indic languages.
\newblock In \emph{Conference on Empirical Methods in Natural Language
  Processing}.

\bibitem[{Lai et~al.(2023)Lai, Ngo, Veyseh, Man, Dernoncourt, Bui, and
  Nguyen}]{Lai2023ChatGPTBE}
Viet~Dac Lai, Nghia~Trung Ngo, Amir Pouran~Ben Veyseh, Hieu Man, Franck
  Dernoncourt, Trung Bui, and Thien~Huu Nguyen. 2023.
\newblock \href {http://arxiv.org/abs/2304.05613} {Chatgpt beyond english:
  Towards a comprehensive evaluation of large language models in multilingual
  learning}.

\bibitem[{Laskar et~al.(2023{\natexlab{a}})Laskar, Bari, Rahman, Bhuiyan, Joty,
  and Huang}]{laskar-etal-2023-systematic}
Md~Tahmid~Rahman Laskar, M~Saiful Bari, Mizanur Rahman, Md~Amran~Hossen
  Bhuiyan, Shafiq Joty, and Jimmy Huang. 2023{\natexlab{a}}.
\newblock \href {https://doi.org/10.18653/v1/2023.findings-acl.29} {A
  systematic study and comprehensive evaluation of {C}hat{GPT} on benchmark
  datasets}.
\newblock In \emph{Findings of the Association for Computational Linguistics:
  ACL 2023}, pages 431--469, Toronto, Canada. Association for Computational
  Linguistics.

\bibitem[{Laskar et~al.(2023{\natexlab{b}})Laskar, Fu, Chen, and
  Bhushan~TN}]{laskar-etal-2023-building}
Md~Tahmid~Rahman Laskar, Xue-Yong Fu, Cheng Chen, and Shashi Bhushan~TN.
  2023{\natexlab{b}}.
\newblock \href {https://doi.org/10.18653/v1/2023.emnlp-industry.33} {Building
  real-world meeting summarization systems using large language models: A
  practical perspective}.
\newblock In \emph{Proceedings of the 2023 Conference on Empirical Methods in
  Natural Language Processing: Industry Track}, pages 343--352, Singapore.
  Association for Computational Linguistics.

\bibitem[{Laskar et~al.(2022)Laskar, Hoque, and Huang}]{laskar2022domain}
Md~Tahmid~Rahman Laskar, Enamul Hoque, and Jimmy~Xiangji Huang. 2022.
\newblock Domain adaptation with pre-trained transformers for query-focused
  abstractive text summarization.
\newblock \emph{Computational Linguistics}, 48(2):279--320.

\bibitem[{Li and Flanigan(2023)}]{li2023task}
Changmao Li and Jeffrey Flanigan. 2023.
\newblock \href {http://arxiv.org/abs/2312.16337} {Task contamination: Language
  models may not be few-shot anymore}.

\bibitem[{Liu et~al.(2019)Liu, Ott, Goyal, Du, Joshi, Chen, Levy, Lewis,
  Zettlemoyer, and Stoyanov}]{Liu2019RoBERTaAR}
Yinhan Liu, Myle Ott, Naman Goyal, Jingfei Du, Mandar Joshi, Danqi Chen, Omer
  Levy, Mike Lewis, Luke Zettlemoyer, and Veselin Stoyanov. 2019.
\newblock Roberta: A robustly optimized bert pretraining approach.
\newblock \emph{ArXiv}, abs/1907.11692.

\bibitem[{Nayeem and Chali(2017{\natexlab{a}})}]{nayeem-chali-2017-extract}
Mir~Tafseer Nayeem and Yllias Chali. 2017{\natexlab{a}}.
\newblock \href {https://doi.org/10.18653/v1/W17-2407} {Extract with order for
  coherent multi-document summarization}.
\newblock In \emph{Proceedings of {T}ext{G}raphs-11: the Workshop on
  Graph-based Methods for Natural Language Processing}, pages 51--56,
  Vancouver, Canada. Association for Computational Linguistics.

\bibitem[{Nayeem and Chali(2017{\natexlab{b}})}]{10.1145/3132847.3133106}
Mir~Tafseer Nayeem and Yllias Chali. 2017{\natexlab{b}}.
\newblock \href {https://doi.org/10.1145/3132847.3133106} {Paraphrastic fusion
  for abstractive multi-sentence compression generation}.
\newblock In \emph{Proceedings of the 2017 ACM on Conference on Information and
  Knowledge Management}, CIKM '17, page 2223–2226, New York, NY, USA.
  Association for Computing Machinery.

\bibitem[{Nayeem et~al.(2018)Nayeem, Fuad, and
  Chali}]{nayeem-etal-2018-abstractive}
Mir~Tafseer Nayeem, Tanvir~Ahmed Fuad, and Yllias Chali. 2018.
\newblock \href {https://aclanthology.org/C18-1102} {Abstractive unsupervised
  multi-document summarization using paraphrastic sentence fusion}.
\newblock In \emph{Proceedings of the 27th International Conference on
  Computational Linguistics}, pages 1191--1204, Santa Fe, New Mexico, USA.
  Association for Computational Linguistics.

\bibitem[{Ouyang et~al.(2022)Ouyang, Wu, Jiang, Almeida, Wainwright, Mishkin,
  Zhang, Agarwal, Slama, Ray, Schulman, Hilton, Kelton, Miller, Simens, Askell,
  Welinder, Christiano, Leike, and Lowe}]{Ouyang2022TrainingLM}
Long Ouyang, Jeffrey Wu, Xu~Jiang, Diogo Almeida, Carroll Wainwright, Pamela
  Mishkin, Chong Zhang, Sandhini Agarwal, Katarina Slama, Alex Ray, John
  Schulman, Jacob Hilton, Fraser Kelton, Luke Miller, Maddie Simens, Amanda
  Askell, Peter Welinder, Paul~F Christiano, Jan Leike, and Ryan Lowe. 2022.
\newblock \href
  {https://proceedings.neurips.cc/paper_files/paper/2022/file/b1efde53be364a73914f58805a001731-Paper-Conference.pdf}
  {Training language models to follow instructions with human feedback}.
\newblock In \emph{Advances in Neural Information Processing Systems},
  volume~35, pages 27730--27744. Curran Associates, Inc.

\bibitem[{Qin et~al.(2023)Qin, Zhang, Zhang, Chen, Yasunaga, and
  Yang}]{qin2023chatgpt}
Chengwei Qin, Aston Zhang, Zhuosheng Zhang, Jiaao Chen, Michihiro Yasunaga, and
  Diyi Yang. 2023.
\newblock \href {http://arxiv.org/abs/2302.06476} {Is chatgpt a general-purpose
  natural language processing task solver?}

\bibitem[{Rae et~al.(2022)Rae, Borgeaud, Cai, Millican, Hoffmann, Song,
  Aslanides, Henderson, Ring, Young, Rutherford, Hennigan, Menick, Cassirer,
  Powell, van~den Driessche, Hendricks, Rauh, Huang, Glaese, Welbl, Dathathri,
  Huang, Uesato, Mellor, Higgins, Creswell, McAleese, Wu, Elsen, Jayakumar,
  Buchatskaya, Budden, Sutherland, Simonyan, Paganini, Sifre, Martens, Li,
  Kuncoro, Nematzadeh, Gribovskaya, Donato, Lazaridou, Mensch, Lespiau,
  Tsimpoukelli, Grigorev, Fritz, Sottiaux, Pajarskas, Pohlen, Gong, Toyama,
  de~Masson~d'Autume, Li, Terzi, Mikulik, Babuschkin, Clark, de~Las~Casas, Guy,
  Jones, Bradbury, Johnson, Hechtman, Weidinger, Gabriel, Isaac, Lockhart,
  Osindero, Rimell, Dyer, Vinyals, Ayoub, Stanway, Bennett, Hassabis,
  Kavukcuoglu, and Irving}]{Rae2021ScalingLM}
Jack~W. Rae, Sebastian Borgeaud, Trevor Cai, Katie Millican, Jordan Hoffmann,
  Francis Song, John Aslanides, Sarah Henderson, Roman Ring, Susannah Young,
  Eliza Rutherford, Tom Hennigan, Jacob Menick, Albin Cassirer, Richard Powell,
  George van~den Driessche, Lisa~Anne Hendricks, Maribeth Rauh, Po-Sen Huang,
  Amelia Glaese, Johannes Welbl, Sumanth Dathathri, Saffron Huang, Jonathan
  Uesato, John Mellor, Irina Higgins, Antonia Creswell, Nat McAleese, Amy Wu,
  Erich Elsen, Siddhant Jayakumar, Elena Buchatskaya, David Budden, Esme
  Sutherland, Karen Simonyan, Michela Paganini, Laurent Sifre, Lena Martens,
  Xiang~Lorraine Li, Adhiguna Kuncoro, Aida Nematzadeh, Elena Gribovskaya,
  Domenic Donato, Angeliki Lazaridou, Arthur Mensch, Jean-Baptiste Lespiau,
  Maria Tsimpoukelli, Nikolai Grigorev, Doug Fritz, Thibault Sottiaux, Mantas
  Pajarskas, Toby Pohlen, Zhitao Gong, Daniel Toyama, Cyprien
  de~Masson~d'Autume, Yujia Li, Tayfun Terzi, Vladimir Mikulik, Igor
  Babuschkin, Aidan Clark, Diego de~Las~Casas, Aurelia Guy, Chris Jones, James
  Bradbury, Matthew Johnson, Blake Hechtman, Laura Weidinger, Iason Gabriel,
  William Isaac, Ed~Lockhart, Simon Osindero, Laura Rimell, Chris Dyer, Oriol
  Vinyals, Kareem Ayoub, Jeff Stanway, Lorrayne Bennett, Demis Hassabis, Koray
  Kavukcuoglu, and Geoffrey Irving. 2022.
\newblock \href {http://arxiv.org/abs/2112.11446} {Scaling language models:
  Methods, analysis \& insights from training gopher}.

\bibitem[{Roark et~al.(2020)Roark, Wolf-Sonkin, Kirov, Mielke, Johny,
  Demirsahin, and Hall}]{roark2020dakshina}
Brian Roark, Lawrence Wolf-Sonkin, Christo Kirov, Sabrina~J. Mielke, Cibu
  Johny, Isin Demirsahin, and Keith Hall. 2020.
\newblock \href {https://aclanthology.org/2020.lrec-1.294} {Processing {S}outh
  {A}sian languages written in the {L}atin script: the {D}akshina dataset}.
\newblock In \emph{Proceedings of the Twelfth Language Resources and Evaluation
  Conference}, pages 2413--2423, Marseille, France. European Language Resources
  Association.

\bibitem[{Rogers et~al.(2020)Rogers, Kovaleva, and
  Rumshisky}]{rogers-etal-2020-primer}
Anna Rogers, Olga Kovaleva, and Anna Rumshisky. 2020.
\newblock \href {https://doi.org/10.1162/tacl_a_00349} {A primer in
  {BERT}ology: What we know about how {BERT} works}.
\newblock \emph{Transactions of the Association for Computational Linguistics},
  8:842--866.

\bibitem[{Shoeybi et~al.(2019)Shoeybi, Patwary, Puri, LeGresley, Casper, and
  Catanzaro}]{Shoeybi2019MegatronLMTM}
Mohammad Shoeybi, Mostofa Patwary, Raul Puri, Patrick LeGresley, Jared Casper,
  and Bryan Catanzaro. 2019.
\newblock \href {http://arxiv.org/abs/1909.08053} {Megatron-lm: Training
  multi-billion parameter language models using model parallelism}.
\newblock \emph{CoRR}, abs/1909.08053.

\bibitem[{Touvron et~al.(2023)Touvron, Martin, Stone, Albert, Almahairi,
  Babaei, Bashlykov, Batra, Bhargava, Bhosale et~al.}]{touvron2023llama2}
Hugo Touvron, Louis Martin, Kevin Stone, Peter Albert, Amjad Almahairi, Yasmine
  Babaei, Nikolay Bashlykov, Soumya Batra, Prajjwal Bhargava, Shruti Bhosale,
  et~al. 2023.
\newblock Llama 2: Open foundation and fine-tuned chat models.
\newblock \emph{arXiv preprint arXiv:2307.09288}.

\bibitem[{Wikipedia(2023{\natexlab{a}})}]{bengali-language-article}
Wikipedia. 2023{\natexlab{a}}.
\newblock \href {https://en.wikipedia.org/wiki/Bengali_language} {Bengali
  language}.
\newblock Accessed: May 23, 2023.

\bibitem[{Wikipedia(2023{\natexlab{b}})}]{indic-languages-article}
Wikipedia. 2023{\natexlab{b}}.
\newblock \href {https://wiki.apertium.org/wiki/Indic_languages} {Indic
  languages}.
\newblock Accessed: May 23, 2023.

\bibitem[{Williams et~al.(2018)Williams, Nangia, and
  Bowman}]{williams-etal-2018-squad2}
Adina Williams, Nikita Nangia, and Samuel Bowman. 2018.
\newblock \href {https://doi.org/10.18653/v1/N18-1101} {A broad-coverage
  challenge corpus for sentence understanding through inference}.
\newblock In \emph{Proceedings of the 2018 Conference of the North {A}merican
  Chapter of the Association for Computational Linguistics: Human Language
  Technologies, Volume 1 (Long Papers)}, pages 1112--1122, New Orleans,
  Louisiana. Association for Computational Linguistics.

\bibitem[{Zhang et~al.(2022)Zhang, Roller, Goyal, Artetxe, Chen, Chen, Dewan,
  Diab, Li, Lin, Mihaylov, Ott, Shleifer, Shuster, Simig, Koura, Sridhar, Wang,
  and Zettlemoyer}]{Zhang2022OPTOP}
Susan Zhang, Stephen Roller, Naman Goyal, Mikel Artetxe, Moya Chen, Shuohui
  Chen, Christopher Dewan, Mona Diab, Xian Li, Xi~Victoria Lin, Todor Mihaylov,
  Myle Ott, Sam Shleifer, Kurt Shuster, Daniel Simig, Punit~Singh Koura, Anjali
  Sridhar, Tianlu Wang, and Luke Zettlemoyer. 2022.
\newblock \href {http://arxiv.org/abs/2205.01068} {Opt: Open pre-trained
  transformer language models}.

\end{thebibliography}


\end{document}